\documentclass[10pt,twocolumn,letterpaper]{article}

\usepackage[pagenumbers]{iccv} 

\definecolor{iccvblue}{rgb}{0.21,0.49,0.74}

\usepackage{hyperref}
\hypersetup{
    pagebackref,
    breaklinks,
    colorlinks,
    allcolors=iccvblue
}

\usepackage{url}
\usepackage{graphicx}
\usepackage{booktabs}
\usepackage{epsfig}
\usepackage{amsmath,amssymb}
\usepackage{algorithm}
\usepackage{algpseudocode}
\usepackage{dsfont}
\usepackage{wrapfig}
\usepackage{tikz}
\usepackage{comment}
\usepackage{color}  
\usepackage{tabularx}
\usepackage{multirow}
\usepackage{makecell}
\usepackage{enumitem}
\setitemize{noitemsep,topsep=0pt,parsep=0pt,partopsep=0pt}
\usepackage{caption}
\usepackage{bbding}
\usepackage{float}

\def\paperID{10317} 
\def\confName{ICCV}
\def\confYear{2025}

\title{MaskSAM: Auto-prompt SAM with Mask Classification \\ for Volumetric Medical Image Segmentation}

\author{Bin Xie$^1$, Hao Tang$^2$, Bin Duan$^1$, Dawen Cai$^4$, Yan Yan$^3$, and Gady Agam$^1$ \\ \\ 
$^1$Department of Computer Science, Illinois Institute of Technology, USA \\
$^2$School of Computer Science, Peking University, China \\
$^3$Department of Computer Science, University of Illinois Chicago, USA \\
$^4$Department of Cell and Developmental Biology, University of Michigan, USA \\
{\tt\small {bxie9@hawk.iit.edu, haotang@pku.edu.cn, bduan2@hawk.iit.edu, } } \\
{\tt\small { dwcai@umich.edu, yyan55@uic.edu, agam@iit.edu} } 
}

\begin{document}
\maketitle

\begin{abstract}
Segment Anything Model~(SAM), a prompt-driven foundation model for natural image segmentation, has demonstrated impressive zero-shot performance. However, SAM does not work when directly applied to medical image segmentation, since SAM lacks the ability to predict semantic labels, requires additional prompts, and presents suboptimal performance. Following the above issues, we propose MaskSAM, a novel mask classification prompt-free SAM adaptation framework for medical image segmentation. We design a prompt generator combined with the image encoder in SAM to generate a set of auxiliary classifier tokens, auxiliary binary masks, and auxiliary bounding boxes. Each pair of auxiliary mask and box prompts can solve the requirements of extra prompts. The semantic label prediction can be addressed by the sum of the auxiliary classifier tokens and the learnable global classifier tokens in the mask decoder of SAM. Meanwhile, we design a 3D depth-convolution adapter for image embeddings and a 3D depth-MLP adapter for prompt embeddings to efficient fine-tune SAM. Our method achieves state-of-the-art performance on AMOS2022, 90.52\% Dice, which improved by 2.7\% compared to nnUNet. Our method surpasses nnUNet by 1.7\% on ACDC and 1.0\% on Synapse datasets.

\end{abstract}

\section{Introduction}
\label{sec:intro}

Foundation models~\cite{devlin2018bert, he2022masked}, trained on vast and diverse datasets, have shown impressive capabilities in various tasks~\cite{openai2023gpt, radford2021learning} and are revolutionizing artificial intelligence. The extraordinary zero-shot and few-shot generalization abilities of foundation models derive a wide range of downstream tasks and achieve numerous and remarkable progress. In contrast to the traditional methods of training task-specific models from scratch, the ``pre-training then finetuning'' paradigm has proven pivotal, particularly in the realm of computer vision. Segment Anything Model (SAM)~\cite{kirillov2023segment}, pre-trained over 1~billion masks on 11~million natural images, has been recently proposed as a visual foundation model for prompt-driven image segmentation and has gained significant attention. SAM can generate precise object binary masks based on its impressive zero-shot capabilities. As a very important branch of image segmentation, medical image segmentation has been dominated by deep learning medical segmentation methods~\cite{ronneberger2015u,akkus2017deep,avendi2016combined} for the past few years. The existing deep learning models are often tailored for specific tasks and achieve remarkable progress due to a strong inductive bias consumption. This raises an intriguing question: Can SAM still have the ability to revolutionize the field of medical image segmentation? Or can SAM still achieve high-performance results in medical image segmentation by properly fine-tuning based on SAM's strong zero-shot capabilities in natural image segmentation? 

\begin{figure}[!t]
\centering
\includegraphics[width=.99\linewidth]{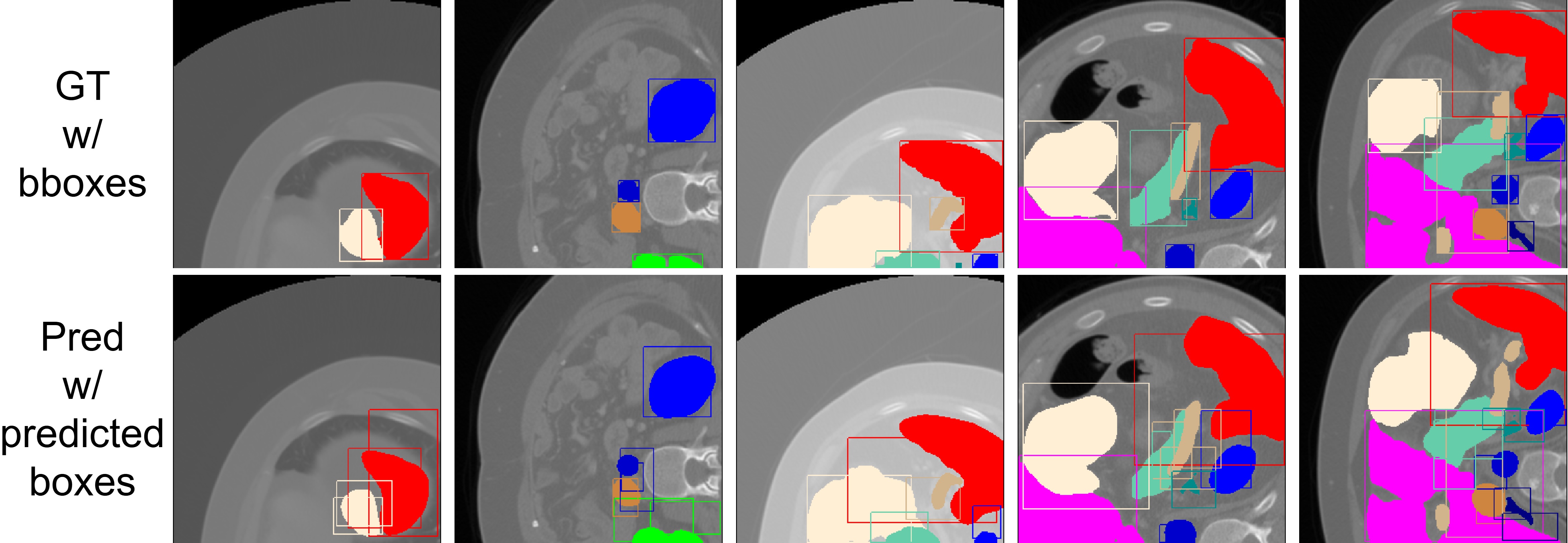}
\caption{Visualization of predicting auxiliary box prompts. Tending to generate multiple prompts to assist in mask prediction.}
\vspace{-0.6cm}
\label{fig:bbox_visual}
\end{figure}

\begin{figure*}[t!]
\centering
    \vspace{-0.6cm}
\includegraphics[width=0.9\linewidth]{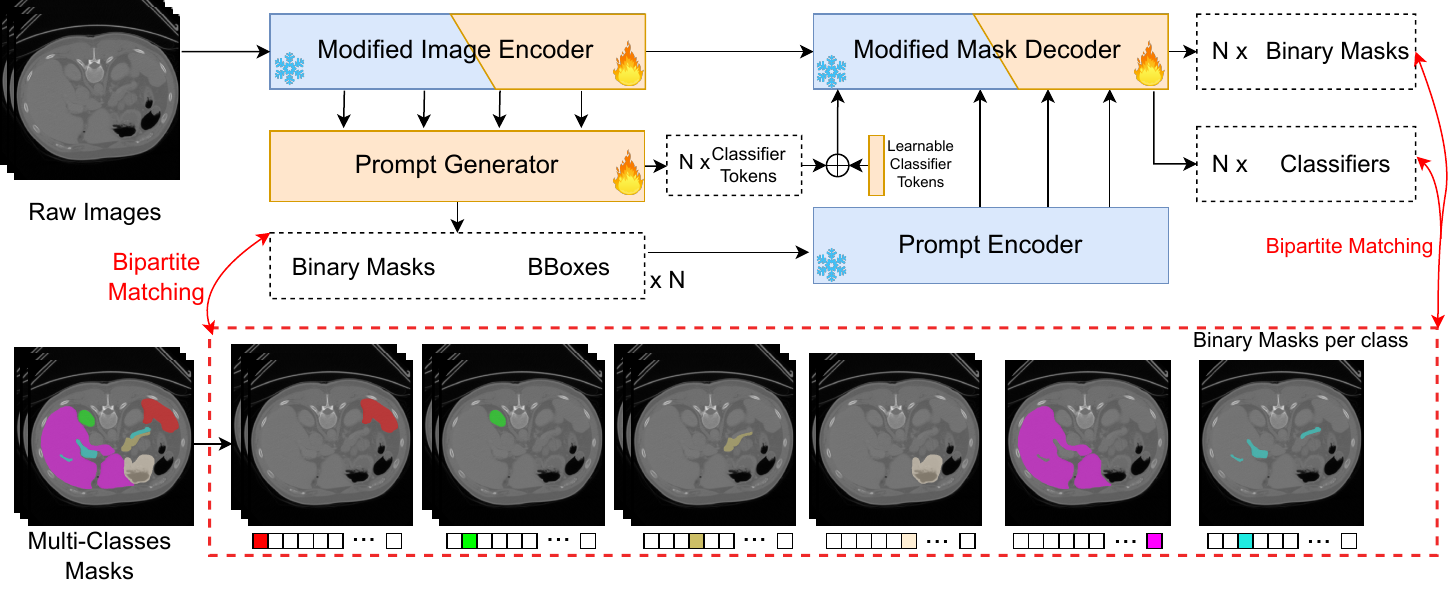}
      \caption{The overview architecture of our proposed MaskSAM.} 
\vspace{-0.6cm}
\label{fig:arch}
\end{figure*}

Over the past year since the publication of SAM, while much research has attempted to adapt SAM to medical image segmentation, few SAM-based models have successfully addressed medical challenges~(\textit{i.e.,} AMOS22~\cite{ji2022amos}), which serve as the most authoritative benchmarks for validating medical image segmentation models. The limitations of existing SAM-based methods in meeting these challenges can be attributed to three key factors: 
1) \underline{inability to predict semantic labels} for each generated binary mask, as SAM exclusively predicts a singular binary mask per prompt without associating semantic labels. Usually, medical images involve multiple labels and each label carries semantic information. These SAM-based methods typically predict binary~(one-class) segmentation. They fall into two categories: the first category does not modify SAM, including directly using SAM or finetuning a part or whole SAM on target datasets, such as MedSAM~\cite{ma2024segment}, Polyp-SAM~\cite{li2024polyp}, and SAM.MD~\cite{wald2023sam}. The methods investigate SAM's performance on medical datasets and need extra prompts. 
Another category modifies SAM but does not achieve semantic label functionality, such as DeSAM~\cite{gao2023desam}, AutoSAM~\cite{shaharabany2023autosam}, and 3DSAM-Adapter~\cite{gong20233dsam}. The methods destroy SAM's zero-shot capabilities so that they can only work on a specific one-class dataset. 
2) \underline{Inabilities to meet extra prompt requirements}. SAM requires users to input precise prompts to segment target regions. These models do not solve the requirement of prompts; they usually use GT to generate prompts during inference, such as Med-SA~\cite{zhang2024segment}, 3DSAM-Adapter~\cite{gong20233dsam}, and SAM-U~\cite{deng2023sam}. These methods cannot participate in any challenge since no labels are provided for the prompts. To compare and demonstrate the effectiveness of our model fairly, we fine-tune it using the same prompts generated from GT as input and achieve excellent performance, 94.70\% Dice and 94.30\% on Synapse and ACDC datasets, which outperform nnUNet by 8.5\% and 2.7\%, respectively. 
3) \underline{Subpar performance}. SAM does not always perform well when applied directly to medical image segmentation, even with the appropriate prompts. Although some SAM-based methods solve semantic label generation and prompt requirements (\textit{i.e.,} SAMed~\cite{zhang2023customized} and SAM3D~\cite{bui2024sam3d}), they exhibit inferior performance. Therefore, the intrinsic issues lie in refining SAM fine-tuning process from natural image segmentation to medical image segmentation resulting in few SAM-based models participating in medical challenges. 



Fortunately, our proposed SAM-based MaskSAM addresses the above issues, exhibits medical segmentation challenges, and achieves state-of-the-art performance for the AMOS22 challenge. To address the extra prompt requirements, we initially designed a prompt-free architecture for SAM.
We observe that the image encoder employs the Vision Transformer~(ViT)~\cite{dosovitskiy2020image} pre-trained with masked auto-encoder~\cite{he2022masked} as the backbone. Capitalizing on ViT's representation capabilities, the image encoder extracts essential features of images with a series of transformer blocks. To address the critical issue of prompt-free usage, we introduce a prompt generator that uses multiple levels of feature maps extracted from the image encoder. The prompt generator generates a set of auxiliary binary masks and bounding boxes as prompts, obviating the need for manual prompts. This innovative approach effectively resolves the requirements for extra prompts.

To enable the prediction of semantic labels, the prompt generator simultaneously produces a set of auxiliary classifier tokens. However, there are no classifier tokens in the mask decoder to output class predictions. Inspired by MaskFormer~\cite{cheng2021per}, we design global learnable classifier tokens, which are summed by auxiliary classifier tokens to each predicted binary mask corresponding to one specific class, in the mask decoder. 

\begin{figure*}[!t]
\centering
  \vspace{-0.6cm}
\includegraphics[width=0.95\textwidth]{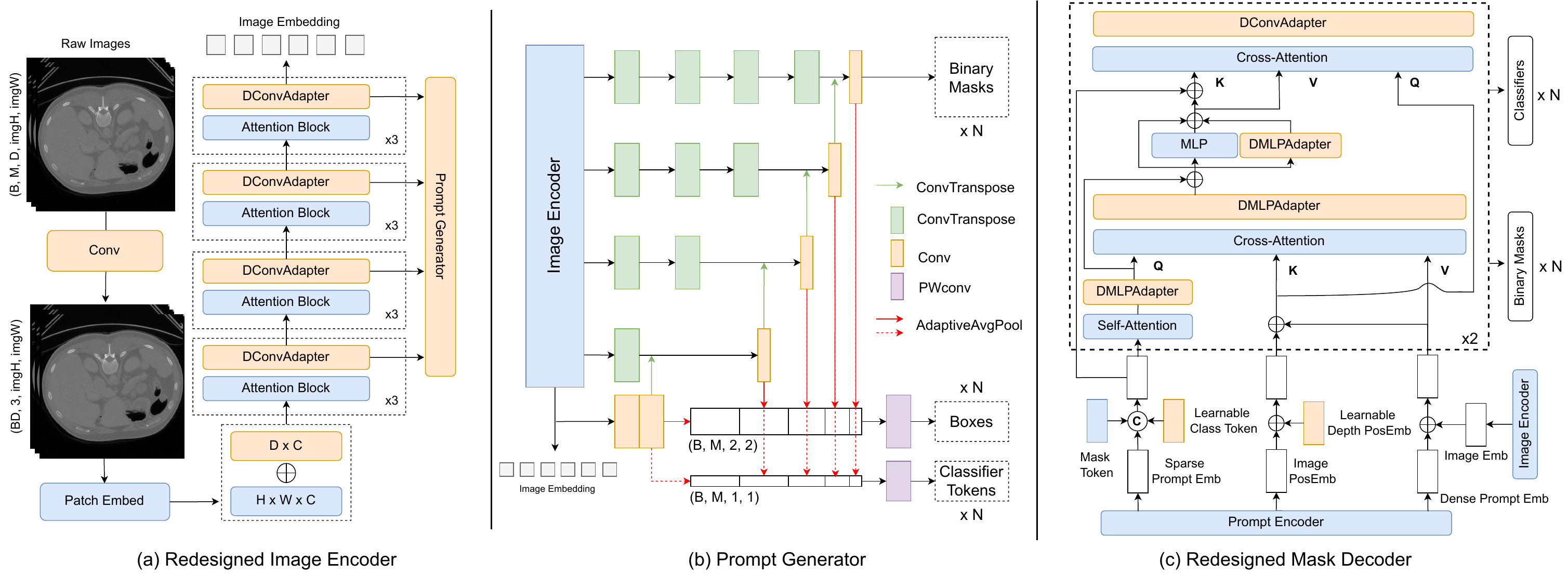}
      \caption{Overview of (a) redesigned image encoder, (b) proposed prompt generator, and (c) redesigned mask decoder. Blue and white boxes are frozen and the rests are tuned.} 
\vspace{-0.6cm}
\label{fig:subarch}
\end{figure*}

We design a dataset mapping pipeline, shown at the bottom of Figure~\ref{fig:arch}, which converts a multi-class mask into a set of binary masks with semantic labels. The dataset mapping pipeline results in various lengths of binary masks of ground truth. Inspired by DETR~\cite{carion2020end} and MaskFormer~\cite{cheng2021per}, the designed prompt generator would generate a large enough number of prompts so that the number is larger than the maximum of class-level binary masks in datasets. Then bipartite matching is utilized between the set of predictions and ground truth segments to select the most matching predicted masks with ground truth segments to calculate losses. 

Despite addressing its functionality issues, SAM does not always perform well when applied directly to medical image segmentation tasks, even with appropriate prompts. Many works~\cite{deng2023segment, hu2023sam, zhou2023can, mohapatra2023sam, roy2023sam, wang2023sam, he2023accuracy} have demonstrated that SAM is imperfect or even fails when some situations occur, such as weak boundaries, low-contrast, and smaller and irregular shapes, which is consistent with other investigations~\cite{ji2023sam, ji2023segment}. Therefore, fine-tuning SAM for medical image segmentation tasks is the main direction. However, fine-tuning the large model of SAM consumes huge computation resources. Many efficient fine-tuning works~\cite {ma2023segment, wu2023medical, li2023auto, gong20233dsam} demonstrate the effectiveness of SAM and the feasibility of efficient fine-tuning by inserting lightweight adapters~\cite{houlsby2019parameter} in medical image segmentation. 
In this paper, we employ the lightweight adapter for efficient fine-tuning. However, the above works usually abandon the prompt encoder or mask decoder to avoid the requirements of additional prompts provided, which would destroy the consistent system of SAM and abandon the strong prompt encoder and mask decoder, which are trained via large-scale datasets. The primary challenge lies in modifying the structure to preserve the inherent capabilities of SAM. Therefore, we keep all structures, freeze all weights, and only insert designed blocks into SAM to adapt. In this way, we retain the zero-shot SAM's capabilities, adapting to medical image segmentation.  

\begin{figure}[!t]
  \vspace{-0.2cm}
  \begin{center}
    \includegraphics[width=0.8\linewidth]{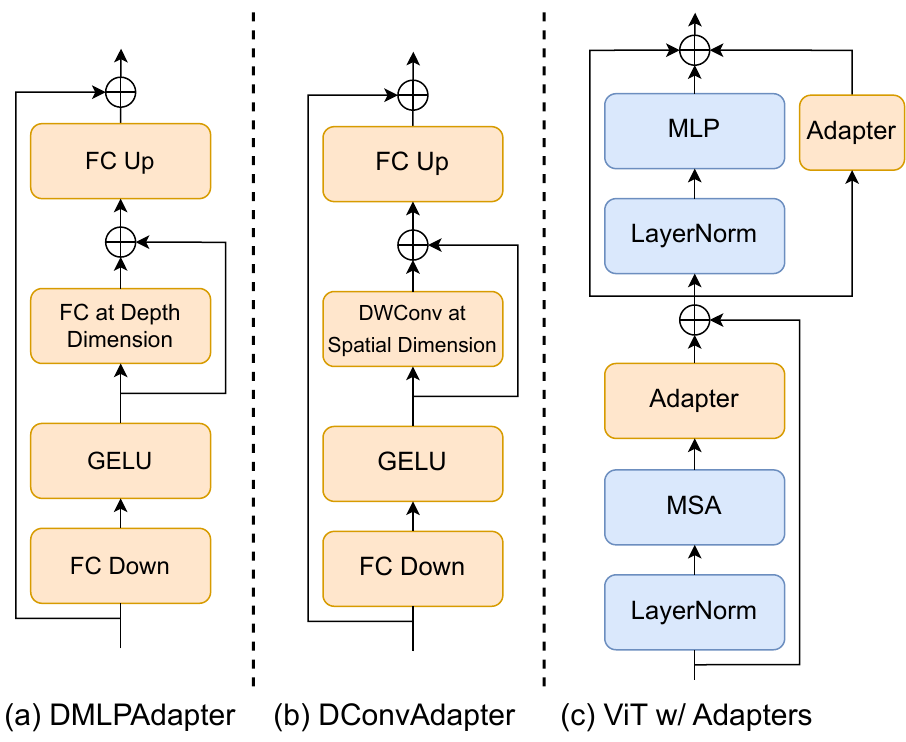}
  \end{center}
  \caption{The proposed adapters.}
  \vspace{-0.6cm}
\label{fig:adapters}
\end{figure}

\begin{figure*}[!t]
\centering
    \vspace{-0.6cm}
\includegraphics[width=0.88\textwidth]{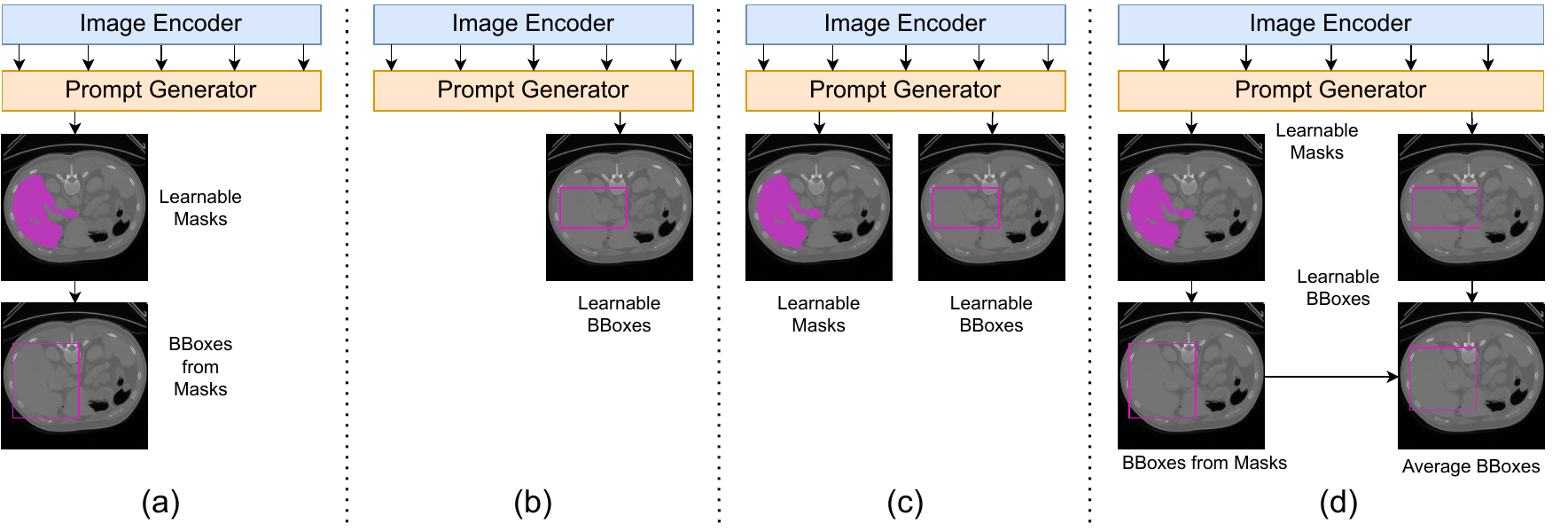}
      \caption{(a) A prompt generator with learnable masks. (b) A prompt generator with learnable boxes. (c) A prompt generator with learnable masks and learnable boxes. (d) A prompt generator with learnable masks and average boxes.}
\vspace{-0.6cm}
\label{fig:prompt_generator}
\end{figure*}

Unlike classic 2D natural images, many medical scans are 3D volumes with an extra depth dimension, such as MRI and CT. To learn extra depth information, we include learnable layers for the extra depth dimension in the lightweight adapters. 
In SAM, the image encoder and mask decoder contain transformer blocks into which we can insert adapters. The mask decoder contains two types of attention blocks for prompt embeddings and image embeddings, respectively. The original adapter only processes the last dimension, the channel dimension, which cannot learn the information among tokens. However, image embeddings contain spatial information, which is important to let our model understand spatial relationships. we design a 3D depth-convolution adapter~(DConvAdapter) that adds a 3D depth-wise convolution layer in the middle of the original adapter with a skip connection for all attention blocks processing image embeddings in the mask decoder. For the rest of the attention blocks for prompt embeddings, we only need to involve a learnable block at depth dimension, since the prompt embeddings do not have any spatial relationship. Therefore, we design a 3D depth-MLP adapter~(DMLPAdapter) that adds an invert-bottleneck architecture that consists of two FC layers and an activation layer processing on depth dimension in the middle of the original adapter with a skip connection, which can learn the extra depth information. Both designed adapters are shown in Figure~\ref{fig:adapters}. Since the image encoder only contains the attention blocks for image embeddings, we insert a DConvAdapter for each transformer block in the image encoder.

In summary, our contributions are as follows: 

\begin{itemize}[leftmargin=*]
\item We propose a novel mask classification prompt-free SAM framework, named MaskSAM, that remains all SAM's structure and weights for medical image segmentation;
\item We propose a novel prompt generator that uses multiple levels of feature maps from the image encoder to generate auxiliary masks and bounding boxes as prompts to solve the requirements of extra prompts and generate auxiliary classifier tokens summed by learnable global classifier tokens within the mask decoder to enable the functionality to predict semantic labels for predicted binary masks;
\item We design a 3D depth-convolution adapter~(DConvAdapter) for image embeddings and a 3D depth-MLP adapter~(DMLPAdapter) for prompt embeddings and inject one of them into each transformer block in the image encoder and mask decoder to enable pre-trained 2D SAM models to extract 3D information and adapt to 3D medical image segmentation;
\item Extensive experiments on three challenging AMOS~\cite{ji2022amos}, ACDC \cite{bernard2018deep} and Synapse \cite{landman2015miccai} datasets demonstrate that MaskSAM achieves state-of-the-art performance, surpassing nnUNet by 2.7\%, 1.7\%, and 1.0\% on AMOS2022, ACDC, and Synapse datasets, respectively. 
\end{itemize}

\section{The Proposed Method}
\label{sec:method}

In this section, we first review SAM. Then, we introduce the whole structure of our proposed MaskSAM. Finally, we describe each component of MaskSAM. 

\noindent \textbf{SAM Preliminaries}
SAM is a prompt-driven foundation model for natural image segmentation, which is trained on the large-scale SA-1B dataset of 1B masks and 11M images. 
The architecture of SAM contains three main components: the image encoder that employs Vision Transformer as the backbone, the prompt encoder that embeds various types of prompts including points, boxes, or texts, and the lightweight mask decoder to generate masks based on the image embedding, prompt embedding, image positional embedding, and output tokens. When utilized to segment a provided 2D image, SAM requires proper prompts, such as points or boxes. Subsequently, SAM generates a singular binary mask for each prompt without any associated semantic labels. However, medical segmentation tasks often involve multiple objects with distinct semantic labels within a given image. 



\subsection{Overview of the Proposed MaskSAM}
In this and the following sections, we introduce the whole pipeline and each component of MaskSAM shown in Figure~\ref{fig:arch} and Figure~\ref{fig:subarch}. Our MaskSAM retains all structures of SAM and only inserts designed blocks to adapt the original SAM from 2D natural images to 3D medical images. Therefore, MaskSAM contains the modified image encoder, the designed generator prompt, the original prompt encoder, and the modified mask decoder. Meanwhile, we design a dataset mapping from multi-class labels to binary masks for each class with semantic labels, which ensures that each predicted binary mask is dedicated to a single class.

\begin{table*}[!t]\small
    \setlength{\tabcolsep}{3pt}
    \centering
    \vspace{-0.6cm}
    \resizebox{1\linewidth}{!}{ 
    \begin{tabular}{@{}c|c|l|ccccccccccccccc|c@{}}
    \toprule
    
    Semantic labels & Prompts & Method & Spl. & R.Kd & L.Kd & GB & Eso. & Liver & Stom. & Aorta & IVC  & Panc. & RAG & LAG & Duo. & Blad. &  Pros. & DSC \\
    \midrule
    
    & & TransBTS \cite{wang2021transbts} & 0.885 & 0.931 & 0.916 & 0.817 & 0.744 & 0.969 & 0.837 & 0.914 & 0.855 & 0.724 & 0.630 & 0.566 & 0.704 & 0.741 & 0.650 & 0.792 \\
    & & UNETR \cite{hatamizadeh2022unetr} & 0.926 & 0.936 & 0.918 & 0.785 & 0.702 & 0.969 & 0.788 & 0.893 & 0.828 & 0.732 & 0.717 & 0.554 & 0.658 & 0.683 & 0.722 & 0.762 \\
    \CheckmarkBold & -- & nnFormer \cite{zhou2021nnformer} & 0.935 & 0.904 & 0.887 & 0.836 & 0.712 & 0.964 & 0.798 & 0.901 & 0.821 & 0.734 & 0.665 & 0.587 & 0.641 & 0.744 & 0.714 & 0.790 \\
    & & SwinUNETR \cite{hatamizadeh2021swin} & 0.959 & 0.960 & 0.949 & \textbf{0.894} & 0.827 & 0.979 & 0.899 & 0.944 & 0.899 & 0.828 & 0.791 & 0.745 & 0.817 & 0.875 & 0.841 & 0.880 \\
    & & nn-UNet~\cite{isensee2019automated} & \textbf{0.965} & 0.959 & 0.951 & 0.889 & 0.820 & 0.980 & 0.890 & 0.948 & 0.901 & 0.821 & 0.785 & 0.739 & 0.806 & 0.869 & 0.839 & 0.878 \\
    \hline
    \XSolidBrush & nnUNet & SAM~\cite{kirillov2023segment} bbox   & 0.679 & 0.741 & 0.640 & 0.168 & 0.443 & 0.773 & 0.671 & 0.651 & 0.554 & 0.434 & 0.232 & 0.324 & 0.444 & 0.698 & 0.602 & 0.538
    \\ 
    \XSolidBrush & nnUNet & SAM 2~\cite{ravi2024sam} bbox & 0.784 & 0.817 & 0.819 & 0.664 & 0.734 & 0.780 & 0.697 & 0.793 & 0.739 & 0.536 & 0.457 & 0.604 & 0.563 & 0.744 & 0.691 & 0.695 \\
    \XSolidBrush & nnUNet & MedSAM~\cite{ma2024segment} bbox   & 0.714 & 0.811 & 0.702 & 0.193 & 0.469 & 0.759 & 0.725 & 0.701 & 0.681 & 0.434 & 0.365 & 0.412 & 0.462 & 0.783 & 0.758 & 0.600
    \\ \hline
    \CheckmarkBold & No needs & SAMed~\cite{zhang2023customized}  & 0.849 & 0.857 & 0.830 & 0.573 & 0.733 & 0.894 & 0.816 & 0.855 & 0.784 & 0.727 & 0.622 & 0.683 & 0.701 & 0.844 & 0.819 & 0.772
    \\
    \CheckmarkBold & No needs & SAM3D~\cite{bui2024sam3d}  & 0.796 & 0.863 & 0.871 & 0.428 & 0.711 & 0.908 & 0.833 & 0.878 & 0.749 & 0.699 & 0.564 & 0.607 & 0.635 & 0.884 & 0.840 & 0.751
    \\ 
    \hline
    \hline
    \CheckmarkBold & No needs & MaskSAM (Ours)   & 0.963 & \textbf{0.973} & \textbf{0.969} & 0.872 & \textbf{0.876} & \textbf{0.982} & \textbf{0.940} & \textbf{0.962} & \textbf{0.922} & \textbf{0.888} & \textbf{0.794} & \textbf{0.813} & \textbf{0.851} & \textbf{0.920} & \textbf{0.854} & \textbf{0.905}
    \\
    \bottomrule
    \end{tabular}}
    \caption{The comparison of MaskSAM with SOTA methods on the AMOS testing dataset evaluated by Dice Score. To fair comparison, all results are based on 5-fold cross-validation without any ensembles. ``Semantic labels'' indicate the model's ability for semantic labeling, while ``Prompt'' specifies the source of the prompt. The best results are indicated as in \textbf{bold}.}
    \label{tab:amos}
\vspace{-0.6cm}
\end{table*}

\subsection{Proposed Dataset Mapping}
SAM generates a single binary mask without any semantic label for each prompt, while a typical ground truth of medical images comprises multiple classes. Each cropped patch image as input may contain varied classes. The challenge arises from varying the lengths of binary masks within the ground truth. To accommodate this diversity and inspired by DETR~\cite{carion2020end} and MaskFormer~\cite{cheng2021per}, our model generates a sufficiently large number of binary masks, with each mask dedicated to predicting a single class. And then bipartite matching is utilized between the set of predictions and ground truth segments. Therefore, we design a dataset mapping pipeline, shown at the bottom of Figure~\ref{fig:arch}, which converts a multi-class mask into a set of binary masks with semantic labels per class. 

\subsection{Proposed Prompt Generator}
To solve the requirements of the extra proper prompts, we propose a prompt generator, shown in Figure~\ref{fig:subarch}(b), to generate a set of auxiliary binary masks and bounding boxes as prompts instead of manual prompts for the prompt encoder. We adopt box prompts with mask prompts as our prompt way, since the point prompts would bring instabilities that are harmful to medical segmentation tasks. We utilize ViT's strong representation capabilities to extract multiple levels of feature maps from the image encoder as the input of our proposed prompt generator. From the beginning of the last output of the image encoder, the feature maps are connected with convolution layers, upsampled, and concatenated with the feature maps of the lower level. Following this way, we can obtain feature maps with the same image size as the ground truth. Finally, we utilize a convolutional layer to change the channel size to the \textit{fixed} number of $N$ that is larger than the maximum of object-level binary masks in datasets. Meanwhile, we extract the output of the last convolutional layer at each level, use adaptive average pooling layers to change the spatial dimension to $(2, 2)$ for box queries, concatenate all box queries, and connect with an MLP layer to adjust the channel to $N$. We can obtain $N$ number of learnable binary masks and learnable boxes.

There are many combinations of learnable binary masks and learnable boxes shown in Figure~\ref{fig:prompt_generator}(a)-(d). Figure~\ref{fig:prompt_generator}(a) shows a prompt generator only generates learnable binary masks and uses the binary masks to calculate its bounding boxes. Figure~\ref{fig:prompt_generator}(b) shows a prompt generator only generates boxes as prompts. Figure~\ref{fig:prompt_generator}(c) shows a prompt generator that generates learnable binary masks as mask prompts and learnable boxes as box prompts. Figure~\ref{fig:prompt_generator}(d) shows a prompt generator that generates learnable binary masks as mask prompts and learnable boxes. We average the bounding boxes calculated from binary masks and the learnable boxes as the final box prompts. Through a series of experiments, we found that the best way is Figure~\ref{fig:prompt_generator}(d) since it can involve more information and more robustness. 

To solve the inabilities of semantic label predictions, the prompt generator simultaneously produces a set of auxiliary classifier tokens, which are the following in the same way as the generation of auxiliary box prompts, except the use of adaptive average pooling layers to change the spatial dimension to $(1, 1)$ for classifier tokens. The auxiliary classifier tokens will be summed up by our designed learnable global classifier tokens within the mask decoder. 

\begin{figure*}[!t]
\centering
    \vspace{-0.6cm}
\includegraphics[width=0.85\linewidth]{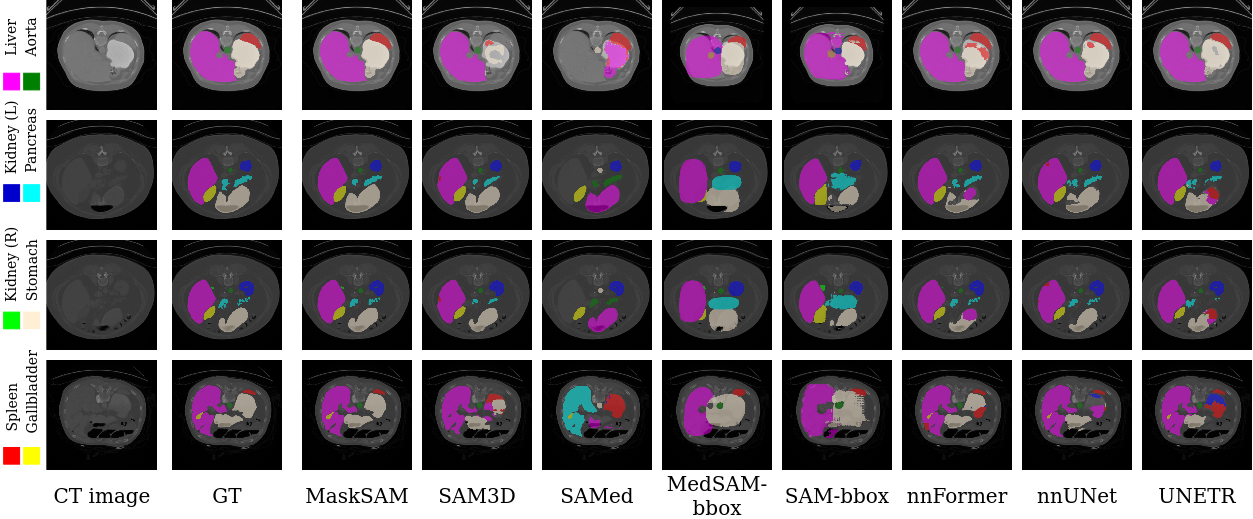}
  \caption{Qualitative comparison on Synapse dataset. MaskSAM is the most precise for each class and has fewer segmentation outliers.}
\vspace{-0.6cm}
\label{fig:sotaSynapse}
\end{figure*}

\subsection{Proposed Adapters}
we adopt the lightweight adapter~\cite{houlsby2019parameter} which is a bottleneck architecture that consists of two fully connected (FC) layers and an activation layer in the middle to modify and inject into each transformer block during fine-tuning. Unlike classic 2D natural images, many medical scans are 3D volumes with extra depth dimensions such as MRI and CT. To learn extra depth information, we involve learnable layers for the extra depth dimension into adapters. In SAM, the image encoder and mask decoder contain transformer blocks that we can insert adapters into. The mask decoder contains two types of attention blocks for prompt embeddings and image embeddings, respectively. The original adapter only processes the last dimension, the channel dimension, which cannot learn the information among the tokens. However, image embeddings contain spatial information, which is important to let our model understand spatial relationships. we design a 3D depth-convolution adapter~(DConvAdapter), shown in Figure~\ref{fig:adapters}(b), that adds a 3D depth-wise convolution layer in the middle of the original adapter with a skip connection for all attention blocks processing image embeddings in the mask decoder. For the rest of the attention blocks for prompt embeddings, we only need to involve a learnable block at depth dimension, since the prompt embeddings do not have any spatial relationship. Therefore, we design a 3D depth-MLP adapter~(DMLPAdapter), shown in Figure~\ref{fig:adapters}(a), that adds an invert-bottleneck architecture that consists of two FC layers and an activation layer processing in the depth dimension in the middle of the original adapter with a skip connection, which can learn additional depth information. Since the image encoder only contains the attention blocks for the image embeddings, we insert the DConvAdapter for each transformer block into the image encoder. Figure~\ref{fig:adapters}(c) shows the way we insert adapters into vision transformers. We insert an adapter behind the multi-head attention block and parallel MLP block.

\subsection{Modified Image Encoder}
Figure~\ref{fig:subarch}(a) illustrates the redesigned image encoder. i) SAM works on natural images that have 3 channels for RGB while medical images have varied modalities as channels. There are gaps between the varied modalities of medical images and the RGB channels of natural images. 
Therefore, we design a sequence of convolutional layers to an invert-bottleneck architecture to learn the adaption from the varied modalities with any size to 3 channels. 
ii) The image encoder includes one positional embedding. To better understand the extra depth information, we can insert a learnable depth positional embedding with the original positional embedding. iii) Since we use the base ViT backbone, it contains 12 attention blocks. We insert our designed DConvAdapter blocks into each attention block. We extract the feature maps of each three attention blocks and the final output of the image encoder for the prompt generator.

\subsection{Modified Mask Decoder}
Figure~\ref{fig:subarch}(c) illustrates the redesigned mask encoder. i) we design learnable global classifier tokens, which are summed by auxiliary classifier tokens generated by the prompt generator, concatenated with sparse prompt embedding, and the original mask tokens to equip our model with the functionality to predict semantic labels for each binary mask. ii) The mask decoder also includes a positional embedding. To learn the extra depth information better, we can insert a learnable depth positional embedding with the original image positional embedding. iii) The mask encoder contains two subsequent transformers. Each transformer first applies self-attention to the prompt embedding. We insert a DMLPAdapter behind the self-attention. Then, a cross-attention block is adopted for tokens attending to image embedding. We insert a DMLPAdapter behind the cross-attention. Next, we insert a DMLPAdapter parallel to an MLP block. Finally, a cross-attention block is utilized for image embedding attending to tokens. We insert a DConvAdapter behind the cross-attention. In this way, our model can learn spatial information with extra depth information for the image embedding and depth information for the prompt embedding. 

\begin{table}[!t]\small
    \setlength{\tabcolsep}{3pt}
    \centering
    \resizebox{0.99\linewidth}{!}{ 
    \begin{tabular}{@{}c|c|l|cccccccc|c@{}}
    \toprule
    Sem. Lab. & Prompts & Method & Aorta & GB  & L.Kd & R.Kd & Liv. & Panc. & Spl. & Stom.  & DSC \\
    \midrule
    & & TransUNet \cite{chen2021transunet}  &  87.23 & 63.16 & 81.87 & 77.02 & 94.08 & 55.86 & 85.08 & 75.62 & 77.48\\
    & & SwinUNet \cite{cao2021swin}   & 85.47 & 66.53 & 83.28 & 79.61 & 94.29 &  56.58 & 90.66 & 76.6 & 79.13\\
    \CheckmarkBold & -- &  UNETR \cite{hatamizadeh2022unetr}   & 89.99 & 60.56 & 85.66 & 84.80 & 94.46 & 59.25 & 87.81 & 73.99 & 79.56\\
    & &  nnUNet \cite{isensee2019automated}  & 92.39 & 71.71 & 86.07 & \textbf{91.46} & 95.84 & 82.92 & 90.31 & 79.01 & 86.21\\
    & &  nnFormer \cite{zhou2021nnformer} & \textbf{92.40} & 70.17 & 86.57 & 86.25 & 96.84 & \textbf{83.35} & 90.51 & 86.83 & 86.57 \\  \hline
    \XSolidBrush & GT & SAM~\cite{kirillov2023segment} bbox   & 60.05 & 24.90 & 68.87 & 54.22 & 76.91 & 45.36 & 69.20 & 67.93 & 58.43 
    \\ 
    \XSolidBrush & GT & MedSAM~\cite{ma2024segment} bbox   & 70.16 & 22.44 & 79.08 & 64.63 & 76.38 & 52.31 & 73.07 & 79.10 & 64.65 \\ 

    \XSolidBrush & GT & MaskSAM (Ours) bbox  & \underline{96.22} & \underline{93.31} & \underline{94.75} & \underline{94.25} & \underline{97.69} & \underline{89.13} & \underline{96.68} & \underline{95.58} & \underline{94.70}
    
    \\ \hline
    
    \CheckmarkBold & No need & SAMed \cite{zhang2023customized} & 87.77 & 69.11 & 80.45 & 79.95 & 94.80 & 72.17 & 88.72 & 82.06 & 81.88 \\ 
    
    \CheckmarkBold & No need & SAMed\_s \cite{zhang2023customized} & 83.62 & 57.11 & 79.63 & 78.92 & 93.98 &  65.66 & 85.81 & 77.49 & 77.78 \\ 
    
    \CheckmarkBold & No need & SAM3D~\cite{bui2024sam3d}  & 89.57 & 49.81 & 86.31 & 85.64 & 95.42 & 69.32 & 84.29 & 76.11 & 79.56
    \\ 
    \hline
    \hline
    \CheckmarkBold & No need & MaskSAM (Ours)   & 91.75 & \textbf{72.20} & \textbf{87.32} & 88.15 & \textbf{97.21} & 79.62 & \textbf{92.47} & \textbf{89.11} & \textbf{87.23}
    \\
    \bottomrule
    \end{tabular}}
    \caption{
    Comparisons with SOTA methods on Synapse.}
    \vspace{-0.6cm}
    \label{tab:sotaSynapse}
\end{table}

\subsection{Losses and Matching}
Following~\cite{cheng2021per, carion2020end}, we build an auxiliary loss, which consists of a combination of binary cross-entropy and dice loss over auxiliary binary mask prediction~($\mathcal{L}_{mask}^{\text{aux}}$) and a combination of $L_1$ loss and generalized IoU loss~\cite{rezatofighi2019generalized} over bounding box predictions~($\mathcal{L}_{box}$), for our prompt generator. Meanwhile, we build a loss, which consists of the standard classification CE-loss on class predictions and a combination of binary cross-entropy and dice loss on final binary mask predictions~($\mathcal{L}_{mask}^{\text{final}}$), for the final output of our MaskSAM. To find the lowest cost assignment, we use bipartite matching~\cite{cheng2021per, carion2020end} between the ground truths and the combination of the set auxiliary predictions and the set of final predictions. Then, our model selects the same indexes from $N$ auxiliary binary masks and $N$ final binary masks by bipartite matching. Finally, we use the indexes to obtain specific predictions to calculate losses with the ground truths. 

Specifically, the desired final output $z = \{ (p_i, m_i)\}_{i=1}^N$ contains $N$ pairs of binary masks $\{m_i^{\text{final}} | m_i^{\text{final}} \in [0, 1]^{H\times W}\}_{i=1}^N$ with classes of probability distribution $p_i \in \Delta ^{K+1}$, which contains $K$ category labels with an auxiliary ''no object`` label~($\varnothing$). Meanwhile, our model produces $N$ pairs of auxiliary boxes and masks, $z_{\text{aux}} = \{ (b_i^{\text{aux}}, m_i^{\text{aux}}) | b_i^{\text{aux}} \in [0,1]^4, m_i^{\text{aux}} \in \{0, 1\}^{H\times W}\}^{N}_{i=1}$. Additionally, the set of $N^{gt}$ ground truth segments $z^{gt} = \{ 
(c_i^{gt}, b_i^{gt}, m_i^{gt}) | c_i^{gt} \in \{1, ...,K \}, b_i^{gt} \in [0,1]^4, m_i^{gt} \in \{0, 1\}^{H\times W}\}^{N^{gt}}_{i=1}$ is required. Since we set $N \ge N_{gt}$ and pad the set of ground truth labels with ``no object token'' $\varnothing$ to allow one-to-one matching. 
To train the model parameters, given a matching $\sigma$, the main loss $\mathcal{L}_{\text{mask-box-cls}}$ is expressed as follows, $\mathcal{L}_{\text{mask-box-cls}} = $
\begin{align}
     & \sum_{j=1}^N [-\log p_{\sigma}(j)(c^{gt}_j) + \mathds{1}_{c^{gt}_j \neq \varnothing} \mathcal{L}_{mask}^{\text{aux}}(m^{\text{aux}}_{\sigma}(j), m_{j}^{gt}) \nonumber \\
    & + \mathds{1}_{c^{gt}_j \neq \varnothing} \mathcal{L}_{box}(b_{\sigma}(j), b_{j}^{gt}) +  \mathds{1}_{c^{gt}_j \neq \varnothing} \mathcal{L}_{mask}^{\text{final}}(m^{\text{final}}_{\sigma}(j), m_{j}^{gt})].  \nonumber 
\label{equa:finalloss}
\end{align}

\section{Experiments}
\label{sec:experiments}

\begin{figure}[!t]
\centering
    \vspace{-0.6cm}
\includegraphics[width=0.99\linewidth]{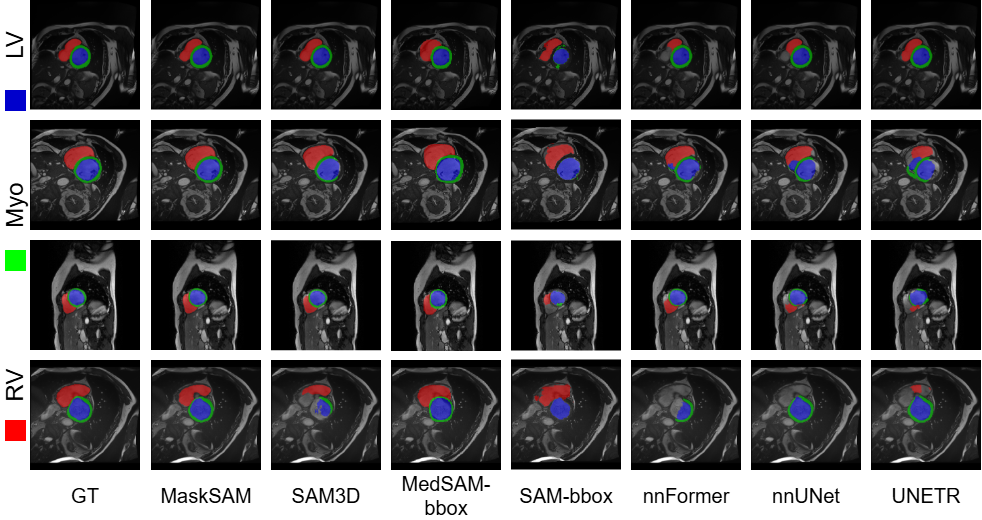}
  \caption{Qualitative comparison on ACDC dataset. MaskSAM is the most precise for each class and has fewer segmentation outliers}
\vspace{-0.6cm}
\label{fig:sotaACDC}
\end{figure}

\noindent \textbf{Datasets and Evaluation Metrics.}
We use three publicly available datasets, AMOS22 Abdominal CT Organ Segmentation~\cite{ji2022amos}, Synapse multiorgan segmentation~\cite{landman2015miccai} and Automatic Cardiac Diagnosis Challenge (ACDC)~\cite{bernard2018deep}. \textbf{(i)} AMOS22 dataset consists of 200 cases of abdominal CT scans with 16 anatomies manually annotated for abdominal multi-organ segmentation. There are 200 testing images, and we evaluate our model on the AMOS22 leaderboard. 
\textbf{(ii)} Synapse dataset consists of 30 cases of abdominal CT scans. Following the split strategies \cite{chen2021transunet},  we use a random split of 18 training cases and 12 cases for validation. We evaluate the model performance via the average Dice score~(DSC) on 8 abdominal organs. \textbf{(iii)} ACDC dataset consists of 100 patients with the labels involving the right ventricle (RV), myocardium (MYO), and left ventricle
(LV). We use a random split of 70 training cases, 10 validation cases, and 20 testing cases. We evaluate the model performance by the average DSC.
In~\cref{tab:amos}, \ref{tab:sotaSynapse}, \ref{tab:sotaACDC}, ``Semantic labels'' indicate the model's ability for semantic labeling, while ``Prompt'' specifies the source of the prompt. Since SAM and MedSAM do not predict semantic labels and need the extra prompts, we utilize GT or the inferred prediction by the pre-trained nnUNet to generate prompts and use the same predictions' labels as the semantic labels. 

\subsection{Comparison with State-of-the-Art Methods}
\noindent \textbf{Results on AMOS 2022 Dataset.}~We present the quantitative outcomes of our experiments on the AMOS 2022 Dataset~\cref{tab:amos}, comparing our proposed MaskSAM against the segmentation methods which are widely used and well-recognized in the community, including convolution-based methods~(nnUNet~\cite{isensee2019automated}), transformer-based methods~(UNETR~\cite{hatamizadeh2022unetr}, SwinUNETR~\cite{hatamizadeh2021swin}, and nnFormer~\cite{zhou2021nnformer}), and SAM-based methods~(SAM~\cite{kirillov2023segment}, MedSAM~\cite{ma2024segment}, SAMed~\cite{zhang2023customized}, and SAM3D~\cite{bui2024sam3d}). To fair comparison, all methods run 5-fold cross-validation without any ensembles. 
We observe that our MaskSAM outperforms all existing methods on most organs, achieving a new state-of-the-art performance in DSC. Meanwhile, the performance of points as the prompts are worse than boxes as the prompts. When utilizing the predictions from nnUNet for bounding box prompts, SAM and MedSAM exhibit decreases of 34\% and 27\%, respectively, compared to nnUNet's accuracy of 87.8\%. These reductions in accuracy indicate negative implications for the results. Specifically, MaskSAM surpasses nnUNet and SwinUNETR by 2.7\% and 2.5\% in DSC, respectively. MaskSAM surpasses SAMed and SAM3D by 13\% and 15\% in DSC, respectively.
In the extremely hard AMOS 2022 dataset, our MaskSAM achieves state-of-the-art performance, which confirms the efficacy of our method. 

\begin{table}[!t]
\centering
 	\vspace{-0.6cm}
          \resizebox{0.95\linewidth}{!}{%

\begin{tabular}{@{}c|c|l|ccc|c@{}}
\toprule
Semantic labels & Prompts & Method & RV $\uparrow$ & Myo $\uparrow$ & LV $\uparrow$ & DSC $\uparrow$ \\ \midrule
& & UNETR \cite{hatamizadeh2022unetr}  & 85.29 & 86.52 & 94.02 & 88.61\\
& &TransUNet \cite{chen2021transunet} & 88.86 & 84.54 & 95.73 & 89.71\\
\CheckmarkBold & -- & SwinUNet \cite{cao2021swin} & 88.55 & 85.62 & 95.83 & 90.00\\ 
& & nnUNet \cite{isensee2019automated} & 90.24 & 89.24 & 95.36 & 91.61\\
& & nnFormer \cite{zhou2021nnformer} & 90.94 & 89.58 & 95.65 & 92.06\\ \hline

\XSolidBrush & GT & SAM~\cite{kirillov2023segment} bbox   & 70.95 & 33.33 & 81.31 & 62.53  
\\ 
\XSolidBrush & GT & MedSAM~\cite{ma2024segment}  bbox & 85.86 & 80.31 & 92.33 & 86.17 \\

\XSolidBrush & GT & MaskSAM bbox  & \underline{94.27} & \underline{91.69} & \underline{96.92} & \underline{94.30}

\\ \hline
\CheckmarkBold & No needs & SAM3D \cite{bui2024sam3d} & 89.44 & 87.12 & 94.67 & 90.41\\
\hline
\hline
\CheckmarkBold & No needs & MaskSAM (Ours) & \textbf{92.30} & \textbf{91.37} & \textbf{96.49} & \textbf{93.39} \\
\bottomrule
\end{tabular}} 
\caption{The comparison of MaskSAM with SOTA methods on ACDC dataset (DSC in \%). The best are in \textbf{bold}.}
\vspace{-0.6cm}
\label{tab:sotaACDC}
\end{table}

\noindent\textbf{Results on ACDC Dataset.}~In~\cref{tab:sotaACDC}, we provide the quantitative experimental results on ACDC dataset. Specifically, we compare the proposed MaskSAM with several leading SAM-based method(\textit{i.e.} SAM3D~\cite{bui2023sam3d}), convolution-based methods (\textit{i.e.}, nnUNet~\cite{isensee2019automated}) and transformer-based methods (\textit{i.e.}, UNETR~\cite{hatamizadeh2022unetr} and nnFormer~\cite{zhou2021nnformer}). The results show that the proposed MaskSAM outperforms various state-of-the-art approaches, surpassing nnFormer by $1.3\%$, $2.3\%$, $1.8\%$, and $0.8\%$ in DSC, RV dice, Myo dice, and LV dice. Meanwhile, our method outperforms the SAM-based methods, SAM~(1 bbox), MedSAM~(1 bbox), and SAM3D, by 31\%, 7\%, 6\%, and 3\%, demonstrating the effectiveness of our method. In~\cref{fig:sotaACDC}, we provide qualitative results compared to several state-of-the-art methods. Qualitative results depicted in~\cref{fig:sotaACDC} further illustrate the model's accuracy across all labels, achieving near-perfect predictions in the challenging, highly saturated dataset. These outcomes affirm the efficacy of our method, as our proposed modules effectively address the limitations of SAM when adapting to medical image segmentation.

\noindent \textbf{Results on Synapse Dataset.}~We present the quantitative results of our experiments in the Synapse dataset in~\cref{tab:sotaSynapse}, comparing our proposed MaskSAM against several leading SAM-based method(\textit{i.e.},~SAMed~\cite{zhang2023customized} and SAM3D~\cite{bui2023sam3d}), convolution-based methods (VNet~\cite{ronneberger2015u} and nnUNet~\cite{isensee2019automated}), transformer-based methods (TransUNet~\cite{chen2021transunet}, SwinUNet~\cite{cao2021swin}, 
and nnFormer~\cite{zhou2021nnformer}). We observe that MaskSAM outperforms all existing methods, achieving a new state-of-the-art performance. Specifically, it surpasses nnFormer by 0.7\% in the DSC for the highly saturated dataset. Meanwhile, our method outperforms the SAM-based methods, SAM~(1 bbox), MedSAM~(1 bbox), SAMed, and SAM3D, by 29\%, 23\%, 6\%, and 8\%, demonstrating the effectiveness of our method. Notably, our model excels in predicting the large-size `Liver,' `Spleen,' and `Stomach' labels, attributed to our innovative DConvAdapter and DMLPAdapter. These adapters enable the learning of more intricate 3D spatial information and adapt 2D SAM to medical image segmentation. In~\cref{fig:sotaSynapse}, we illustrate qualitative results compared to representative methods. These results also demonstrate that our MaskSAM can predict more accurately the `Liver', `Spleen', and `Stomach' labels. In conclusion, the effectiveness of our method is robustly demonstrated by both quantitative and qualitative results.

\subsection{Ablation Study}
\label{sec:ablation}
\noindent\textbf{Baseline Models.} The proposed MaskSAM has 9 baselines (\textit{i.e.}, B1, B2, B3, B4, B5, B6, B7, B8, B9) as shown in~\cref{tab:archAblation}. All baselines contain the entire SAM structure, a prompt generator, and a learnable class token. (i) B1 adopts the prompt generator that only generates learnable binary masks and uses the binary masks as mask prompts to calculate its bounding boxes as box prompts shown in~\cref{fig:prompt_generator}(a). (ii) B2 adopts the prompt generator that only generates learnable boxes as the prompts shown in~\cref{fig:prompt_generator}(b). (iii) B3 adopts the prompt generator that generates learnable binary masks as mask prompts and learnable boxes as box prompts shown in~\cref{fig:prompt_generator}(c). (iv) B4 adopts the prompt generator that generates learnable binary masks as mask prompts and learnable boxes. We average the bounding boxes calculated from binary masks and the learnable boxes as the final box prompts shown in~\cref{fig:prompt_generator}(d). (v) B5 adds depth positional embedding blocks~(DPosEmbed) in the image encoder and mask decoder based on B4. (vi) B6 modifies the vanilla adapter by inserting the invert-bottleneck depth MLPs with a skip connection after the fully-connected layers for upsampling based on B5. (vii) B7 modifies the vanilla adapter by inserting the invert-bottleneck depth MLPs with a skip connection before the fully-connected layers for downsampling based on B5. (viii) B8 replaces the vanilla adapter with our designed DMLPAdapter based on B5. (ix) B9 is our full model, named MaskSAM, illustrated in~\cref{fig:arch}. B9 adopts the DMLPAdapter for prompt embeddings and the DConvAdapter for image embedding based on B8. 

\begin{table}
    \vspace{-0.6cm}
    \centering
    \resizebox{0.99\linewidth}{!}{ 
    \begin{tabular}{@{}cl|c}
    \toprule
    & Method & DSC $\uparrow$   \\ \midrule
    B1 & SAM + MaskPG + vAdapter & 89.53 \\
    B2 & SAM + BBoxPG + vAdapter & 88.78   \\ 
    B3 & SAM + MaskBBoxPG + vAdapter & 90.08   \\
    B4 & SAM + MaskAvgBBoxPG + vAdapter & 91.45   \\ 
    B5 & SAM + MaskAvgBBoxPG + vAdapter + DPosEmbed & 91.61   \\ 
    B6 & SAM + MaskAvgBBoxPG + vAdapter w/ D-MLP after FC-Up + DPosEmbed & 92.88   \\ 
    B7 & SAM + MaskAvgBBoxPG + vAdapter w/ D-MLP before FC-Down + DPosEmbed & 92.93   \\ 
    B8 & SAM + MaskAvgBBoxPG + DMLPAdapter + DPosEmbed & 93.10  \\ \hline
    B9 & Our Full Model~(B8 + DConvAdapter) & \textbf{93.39}  \\
    \bottomrule
    \end{tabular}} 
    \caption{Ablation studies of proposed methods on ACDC. \{\}PG means a prompt generator with different prompts. vAdapter means vanilla adapter. D-MLP means MLP layers on depth dimension. DPosEmbed means depth positional embedding.}
    \label{tab:archAblation}
    \vspace{-0.6cm}
\end{table}

\noindent\textbf{Ablation analysis.} The results of the ablation study are shown in~\cref{tab:archAblation}. When we use our proposed MaskAvgBBoxPG to first generate auxiliary binary masks and auxiliary boxes, then average the bounding boxes calculated from auxiliary binary masks and learnable auxiliary boxes as the final box prompts, the model achieves the best results and improves by 1.9\%, 2.6\% and 1.3\% compared to B1 with a learnable mask prompt generator, B2 with a learnable box prompt generator, and B3 with a learnable mask and a learnable box prompt generator, respectively. The result confirms the effectiveness of the proposed prompt generator. When inserting depth positional embedding~(DPosEmbed) into the image encoder and mask decoder, the performance of B5 improves by more than 0.15\% compared to B4, demonstrating the effectiveness of DPosEmbed blocks. The DMLPAdapter~(B8) which we insert depth MLPs with a skip connection in the middle of the vanilla adapter achieves the best performance and improves by 0.2\% and 0.1\% compared to B6 in which we insert depth MLPs with a skip connection after fully-connected layers for upsampling and B7 in which depth MLPs are inserted with a skip connection after fully-connected layers for downsampling, respectively. The result confirms the effectiveness of the proposed DMLPAdapter. B9 is our full model, MaskSAM, utilizing DMLPAdapter for prompt embeddings and DConvAdapter that we replace the depth MLP with a 3D depth-wise convolution layer from the DMLPAdapter for image embedding based on B8 as shown in~\cref{fig:arch}. Compared to B8, our model brings about 0.3\% improvements. Therefore, the results demonstrate the effectiveness of our proposed MaskSAM. 

\begin{table}[!t]
    \centering
    \vspace{-0.6cm}
    \resizebox{0.99\linewidth}{!}{
    \begin{tabular}{@{}c|c|c|c|c}
    \toprule
    \# Params~(M)  & nnUNet &  MedSAM   & SAMed & Med-SA  \\ \midrule
    Tunable Params & 29M & 91M & 19M & 13M \\ \hline
    Total Params & 29M & 91M & 636M+19M & 636M+13M \\ \hline
    \hline
    \# Params~(M) & SAM3D &  3DSAM-Adapter & AutoProSAM & MaskSAM~(ours)  \\ \midrule
    Tunable Params & 2M & 26M & 27M & 14M \\ \hline
    Total Params & 91M+2M & 91M+26M & 91M+27M & 91M+14M \\ \hline 
    \bottomrule
    \end{tabular}} 
    \caption{Comparisons of tunable and total parameters.}
    \label{tab:params}
    \vspace{-0.6cm}
\end{table}

\subsection{Model Analysis}
\label{sec:ablation}
\noindent\textbf{Efficiency metrics.} \cref{tab:params} compares model efficiency by tunable and total parameters. Our method, with only 14M tunable parameters, effectively adapts the 91M ViT-base SAM model and ensures efficiency.

\noindent \textbf{Box prompt visualization.}~\cref{fig:bbox_visual} shows generated box prompts. For complex objects and images with a few objects, our model tends to generate multiple prompts to assist in mask prediction, as the object queries outnumber the objects in 2D images. The rest of the queries are assigned as no-objects. Additionally, the generated boxes are typically larger than the objects themselves, allowing the model to account for adjacent objects. As a result, the generated auxiliary box prompts are both precise and meaningful.

\section{Conclusion}
\label{sec:conclusion}

In this paper, we propose a mask classification prompt-free SAM adaptation framework for medical image segmentation, named MaskSAM.
By designing a prompt generator combined with the image encoder in SAM to generate a set of auxiliary classifier tokens, binary masks, and bounding boxes to solve the inability to infer semantic labels and its reliance on providing extra prompts.
By inserting one of our redesigned 3D depth-convolution adapter~(DConvAdapter) for image embeddings and 3D depth-MLP adapter~(DMLPAdapter) for prompt embeddings into each transformer block in SAM for efficient fine-tuning, our model enables pre-trained 2D SAM models to extract 3D information and adapt to 3D medical images. Our method achieves state-of-the-art performance on AMOS2022~\cite{ji2022amos}, ACDC~\cite{bernard2018deep}, and Synapse~\cite{landman2015miccai} datasets with 90.52\%, 93.39\%, and 87.23\% Dice, which improved by 2.7\%, 1.7\%, and 1.0\% compared to nnUNet, respectively.

{
    \small
    \bibliographystyle{ieeenat_fullname}
    \bibliography{main}
}

\end{document}